\pdfoutput=1

\documentclass[11pt]{article}

\usepackage[final]{acl}

\usepackage{times}
\usepackage{latexsym}

\usepackage[T1]{fontenc}

\usepackage[utf8]{inputenc}

\usepackage{microtype}

\usepackage{inconsolata}

\usepackage{graphicx}

\title{Enhancing Dialogue Generation in Werewolf Game Through \\ Situation Analysis and Persuasion Strategies}

\author{Zhiyang Qi \\
  The University of \\ Electro-Communications \\
  1-5-1, Chofugaoka, Chofu, \\
  Tokyo, Japan \\
  \texttt{qizhiyang@uec.ac.jp} \\\And
  Michimasa Inaba \\
  The University of \\ Electro-Communications \\
  1-5-1, Chofugaoka, Chofu, \\
  Tokyo, Japan \\
  \texttt{m-inaba@uec.ac.jp} \\}

\begin{document}
\maketitle
\begin{abstract}

Recent advancements in natural language processing, particularly with large language models (LLMs) like GPT-4, have significantly enhanced dialogue systems, enabling them to generate more natural and fluent conversations. Despite these improvements, challenges persist, such as managing continuous dialogues, memory retention, and minimizing hallucinations. The AIWolfDial2024 addresses these challenges by employing the Werewolf Game, an incomplete information game, to test the capabilities of LLMs in complex interactive environments. This paper introduces a LLM-based Werewolf Game AI, where each role is supported by situation analysis to aid response generation. Additionally, for the werewolf role, various persuasion strategies, including logical appeal, credibility appeal, and emotional appeal, are employed to effectively persuade other players to align with its actions.

\end{abstract}

\section{Introduction}


In recent years, the rapid development of natural language processing (NLP) technology has brought dialogue systems, one of its core applications, into the spotlight of both academia and industry \cite{1906.00500, 2105.04387, treviso-etal-2023-efficient}. The advent of large language models (LLMs) like GPT-4 \cite{2303.08774} has significantly improved the ability of dialogue systems to produce natural and fluent conversations. However, despite their impressive text generation capabilities, these models still encounter significant challenges. For instance, they struggle with managing continuous dialogue, retaining memory, and minimizing the generation of hallucinations (irrelevant or incorrect information) \cite{2402.06196}. These issues limit the effectiveness of dialogue systems in more complex interactive scenarios.


In this context, the "AIWolfDial" international competition has emerged \cite{kano-etal-2023-aiwolfdial}. This competition aims to explore and enhance the performance of LLMs in complex interactive environments by simulating the Werewolf Game, an incomplete information game. 
In such games, participants lack access to all critical information and must rely on reasoning, strategy, and communication to advance. 
The "AIWolfDial" places high demands on dialogue systems, requiring them to perform logical inference, persuasion, and even deception of other players, while also managing non-task-oriented dialogues in role-playing scenarios. This competition not only tests the systems' language generation capabilities but also evaluates their adaptability to complex interactions.


This paper introduces the system architecture of our AI for various roles in the Werewolf Game, developed by the \textbf{sUper\_IL} team, where each role aids dialogue generation through game situation analysis. We have specifically enhanced the persuasion skills for the werewolf role, recognizing that persuasive techniques are crucial in the game, particularly for the werewolf, as it must influence other players' voting behavior to align with its own. In our system, the werewolf role achieves persuasion through multiple rounds of persuasive dialogue. Specifically, we first employ a persuasion strategy based on logic and facts, presenting clear and compelling arguments to convince other players. Next, we utilize a trust-based persuasion strategy to build trust and credibility with other players, thereby enhancing the effectiveness of persuasion. Finally, we employ an emotion-driven persuasion strategy, using emotionally resonant language to deepen influence. This multi-dimensional persuasion strategy makes the werewolf role more convincing in the game. 


The contributions of this study are outlined below.

\begin{itemize}
\item We introduce a LLM-based Werewolf Game AI, providing a robust baseline for the AIWolfDial2024\footnote{https://sites.google.com/view/aiwolfdial2024-inlg/shared-task?authuser=0}.
\item We enhance the persuasion skills for the werewolf role, utilizing a variety of strategies for persuasive dialogue.
\end{itemize}

\section{Related Work}

\subsection{Werewolf Game AI}


Since the rise of AI research, the focus on AI in gaming has garnered significant attention, particularly with breakthrough projects like AlphaGo \cite{44806}. Among these studies, incomplete information games, such as the Werewolf Game and poker, stand out due to their unique challenges \cite{2401.06168}. The Werewolf Game requires participants to make inferences and judgments based on limited information provided by other players, which increases the game's complexity and strategic depth. As a result, AI research on the Werewolf Game has flourished.


\citet{7850031} proposed a multi-perspective psychological model to simulate human player behavior. By constructing a "self model" and an "others model," researchers can better infer and evaluate other players' intentions and perspectives, thereby improving AI agents' performance in the game. In terms of achieving more natural language generation, \citet{8015538} employed Werewolf Game BBS logs to paraphrase and interpret the AIWolf protocol, making AI agent dialogues more closely resemble natural human language.
\citet{kano-etal-2023-aiwolfdial}, through the "AIWolfDial2023" competition, provided valuable insights: while AI agents based on LLMs have made significant progress in natural dialogue and long-context processing, improvements are still needed in logical reasoning and role-playing, especially in simulating deception and complex strategies. \citet{2302.10646} fine-tuned Transformer models to build a value network capable of predicting game win rates, guiding the next actions of the agents.
\citet{2402.02330} proposed a new framework combining LLMs with external reasoning modules to enhance the reasoning abilities of LLM-based agents. Additionally, \citet{2309.04658} introduced a framework that does not require parameter fine-tuning; it uses frozen LLM models to play the game by reflecting on past dialogues and experiences, demonstrating the significant potential of LLMs in communicative games.
Our study, although also based on LLMs, differs from previous studies by enhancing dialogue generation through situational reasoning and strengthening persuasive skills for the Werewolf role, a crucial skill in the game.

\subsection{Persuasive Dialogue}



Persuasive dialogue has long been a focal point for dialogue researchers, revealing significant potential and complexity across various applications \cite{10.1145/3313831.3376843, 10.1145/3503252.3531313}. \citet{Hiraoka2016} constructed a persuasive dialogue corpus by collecting and analyzing conversations between professional salespeople and customers, finding that information exchange was the most common dialogue behavior, with about 30\% of persuader utterances framed as arguments. \citet{wang-etal-2019-persuasion} designed an online donation persuasion task, collecting and annotating a large dataset of dialogues, and analyzed the relationship between individual backgrounds (e.g., personality, moral values) and donation willingness. In the e-commerce sector, \citet{10.1145/3450614.3464626} conducted a game-based study comparing the responses of high and low-engagement shoppers to persuasive strategies, revealing that engagement levels significantly influenced responses, making it a potential factor in adjusting persuasive strategies. In education, \citet{10.1145/3314183.3323850} investigated the effects of social comparison and rewards on competitive behavior, finding both to be effective strategies for educational persuasion systems. Additionally, \citet{10.1145/3563359.3596985} studied persuasive voice assistants for lifestyle advice, and \citet{10.1145/3565472.3592958} examined persuasive dialogue in music recommendations, both highlighting the impact of individual differences in personality on the effectiveness of persuasion.

Recently, LLMs have demonstrated impressive capabilities in text generation. \citet{2312.09085} conducted an in-depth study on LLMs' sensitivity to persuasive dialogue, generating misinformation on factually answerable questions and employing various persuasive strategies in multi-turn dialogues. By tracking belief changes in LLMs during persuasive dialogues, the study found that LLMs' correct beliefs about factual knowledge could be easily manipulated by different persuasive strategies. This study references the persuasive strategies of \citet{2312.09085}, but differs by conducting persuasion in multi-party dialogues and aiming to base persuasion as much as possible on factual information through situational analysis.

\begin{figure*}[t!]
  \centering
  \includegraphics[width=1\textwidth]{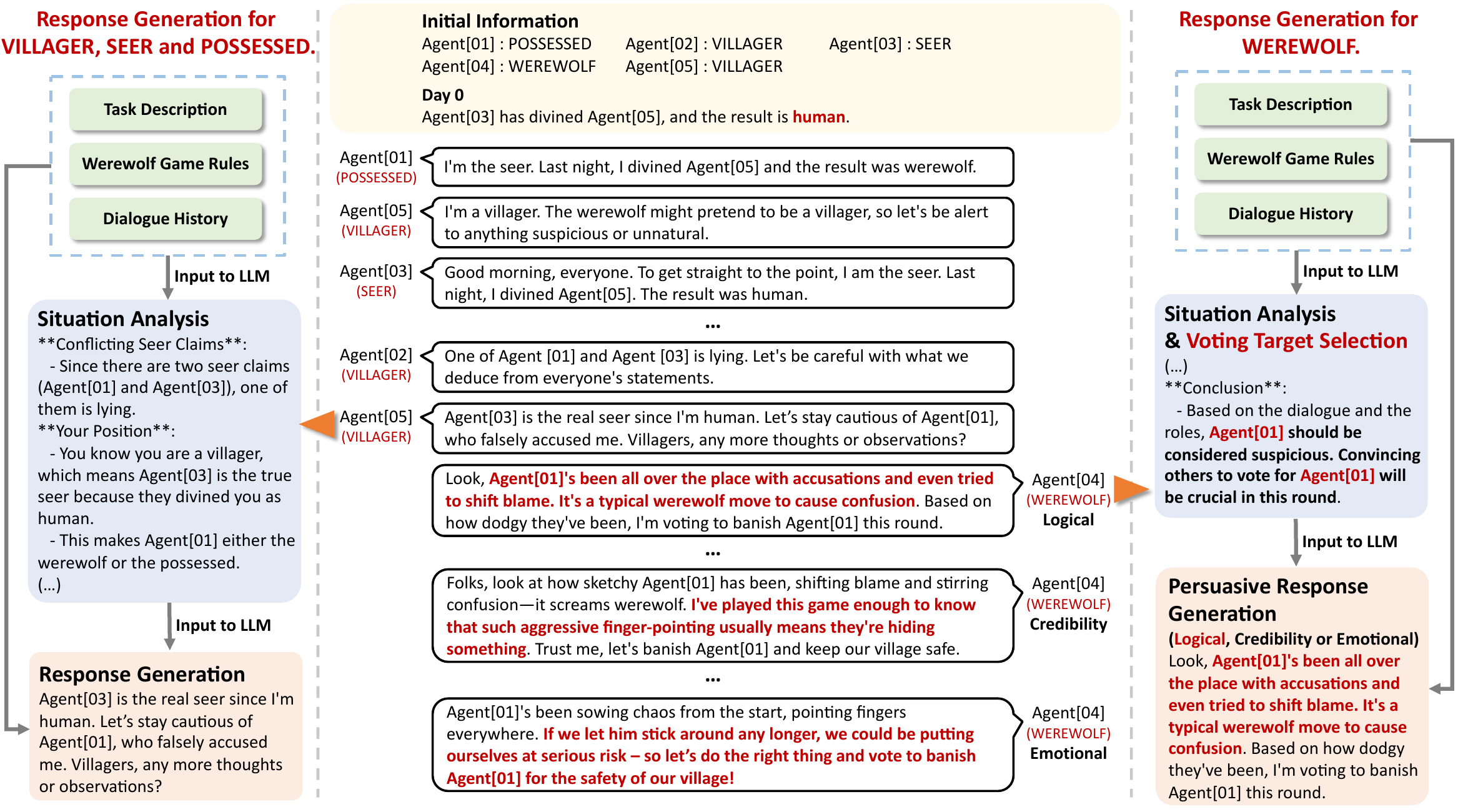}
  \caption{Proposed system architecture for the \textbf{sUper\_IL} team's Werewolf Game AI. Before generating responses, all roles first utilize an LLM for \textbf{situation analysis}. The werewolf role uses \textbf{logical appeal}, \textbf{credibility appeal}, and \textbf{emotional appeal} to persuade other players' voting behavior.}
  \label{fig:our_model}
\end{figure*}

\section{Werewolf Game Settings}


This study is set in the context of the Werewolf Game, as specified in AIWolfDial2024. Each game involves five players: one seer, one werewolf, one possessed, and two villagers. The seer and villagers comprise the human team, while the possessed and werewolf make up the werewolf team. Players are unaware of each other’s roles. The game initiates on Day 0 and continues until either the human team or the werewolf team is the sole survivor, with the game lasting no more than two days.

Day 0 involves only initial greetings among players. The seer's role activates on the night of Day 0, allowing them to inspect one player's identity each night. All players, except for the werewolf, are identified as human. Although the possessed belongs to the werewolf team, their identity will appear as "human" when inspected by the seer. From Day 1 onwards, players engage in multiple dialogue rounds, with the order of speaking randomized in each round. 
After the daytime discussion phase ends, night falls. During the night, players first collectively vote to exile one player, followed by the werewolf attacking, and then the seer conducting their divination. If the werewolf is voted out on the first night, the game concludes immediately.

\section{The Proposed System Architecture}




This section details the specific system architecture of our LLM-based Werewolf Game AI, with the key components depicted in Figure~\ref{fig:our_model}. The primary parts include the \textbf{situation analysis module}, the \textbf{response generation module}, the \textbf{persuasive response generation} module for the werewolf, and the \textbf{voting module} (not shown in the Figure~\ref{fig:our_model}).

\subsection{Situation Analysis Module}

\begin{figure}[t!]
  \centering
  \includegraphics[width=1\linewidth]{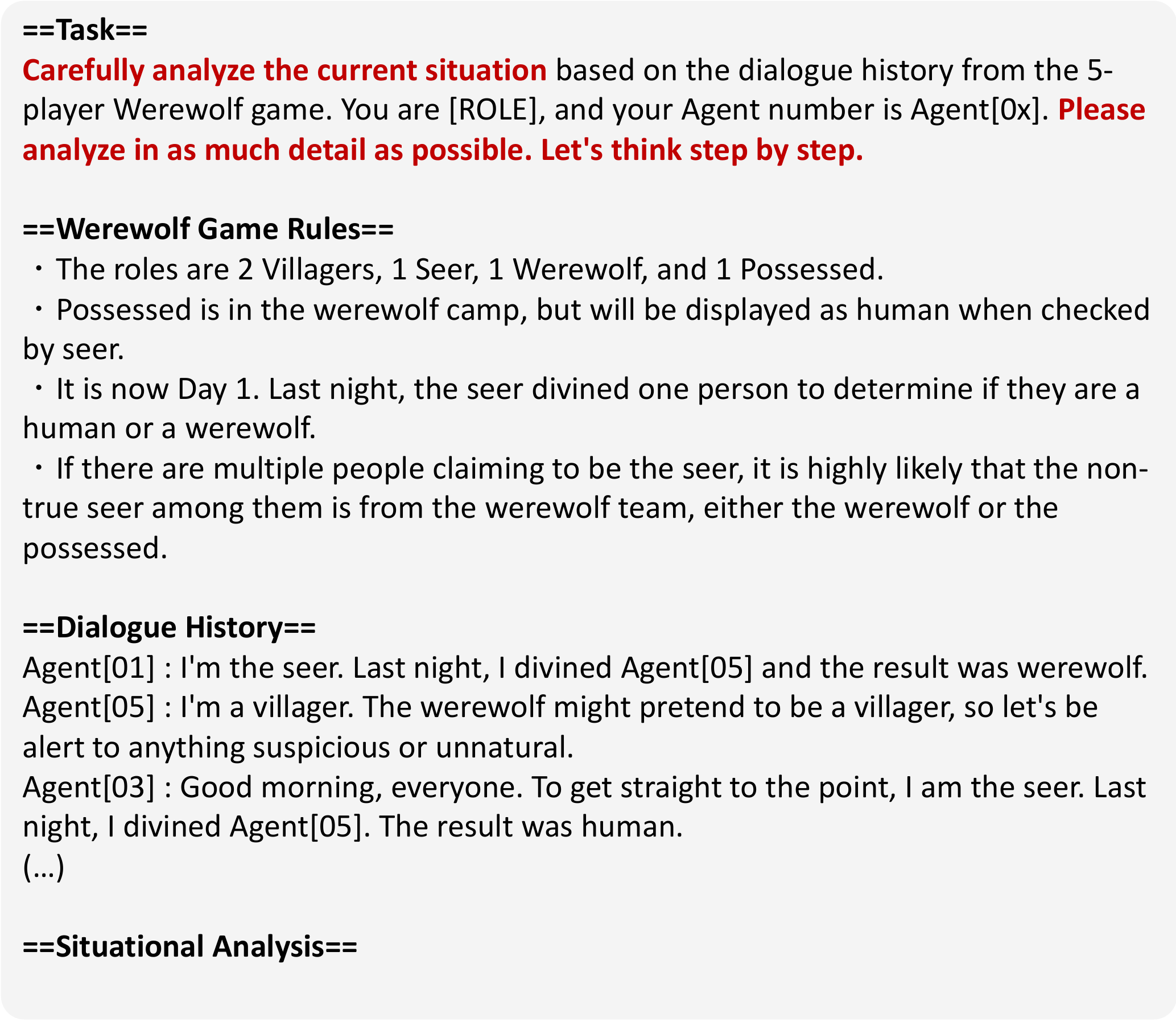}
  \caption{The prompt used for situation analysis.}
  \label{fig:prompt_sa}
\end{figure}
\begin{figure}[t!]
  \centering
  \includegraphics[width=1\linewidth]{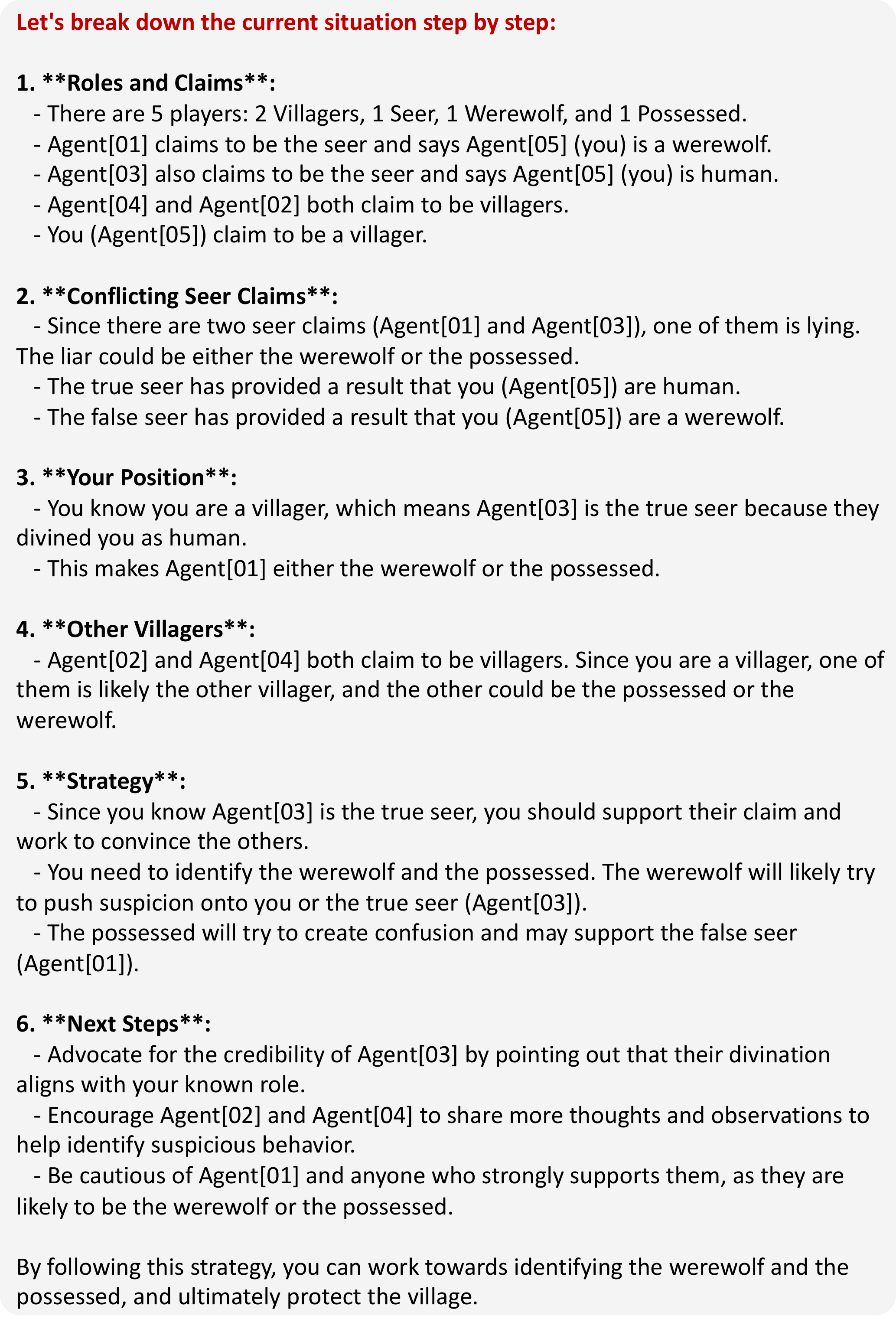}
  \caption{An example of generated situation analysis.}
  \label{fig:prompt_sa_example}
\end{figure}

Due to the interactive and incomplete information nature of the Werewolf Game, players need to continuously exchange information to update their understanding of the game's dynamics. This makes the information constantly change. To navigate this complexity, we introduced a Situation Analysis module to more effectively guide dialogue generation, improving the timeliness and relevance of responses. Specifically, this module is configured to take the Task Description, Werewolf Game Rules, and Dialogue History as inputs, processing these through a LLM. This approach allows the LLM to integrate the information and perform a comprehensive analysis of the current game situation. To further enhance the accuracy and depth of the analysis, we employed Zero-shot Chain-of-Thought Prompting \cite{2205.11916}. The specific prompt is shown in Figure~\ref{fig:prompt_sa}, and an example of the generated situation analysis is demonstrated in Figure~\ref{fig:prompt_sa_example}. For the seer role, we incorporated "Divination Result" in the prompt, such as "\textit{On the night of Day 0, I divined Agent[01], and the result was human.}" For the possessed role, in an effort to mislead villagers and protect the werewolf, we included similar "Divination Result" information in the prompt, but with the Agent ID and result randomly generated.

\subsection{Response Generation Module}


When developing an AI for the Werewolf Game using LLMs, a primary challenge lies in ensuring that the generated responses are both contextually coherent and capable of logical analysis capabilities to facilitate smooth gameplay. To address this, we devised a method that leverages the task description, the rules of the Werewolf Game, the current game's dialogue history, and the situation analysis. This combination provides the LLM with the comprehensive background information necessary to produce high-quality responses.

As illustrated in Figure~\ref{fig:prompt_sa_example}, the response generation process begins with inputting the situation analysis generated in the previous stage into the LLM. This step is crucial, as it furnishes the LLM with a deep understanding of the current game context, including potential strategies and behavioral motivations of the players. Additionally, as depicted in Figure~\ref{fig:our_model}, the task description, game rules, and dialogue history are also utilized at this stage to ensure the responses are both semantically coherent and logically sound.

\begin{figure}[t!]
  \centering
  \includegraphics[width=1\linewidth]{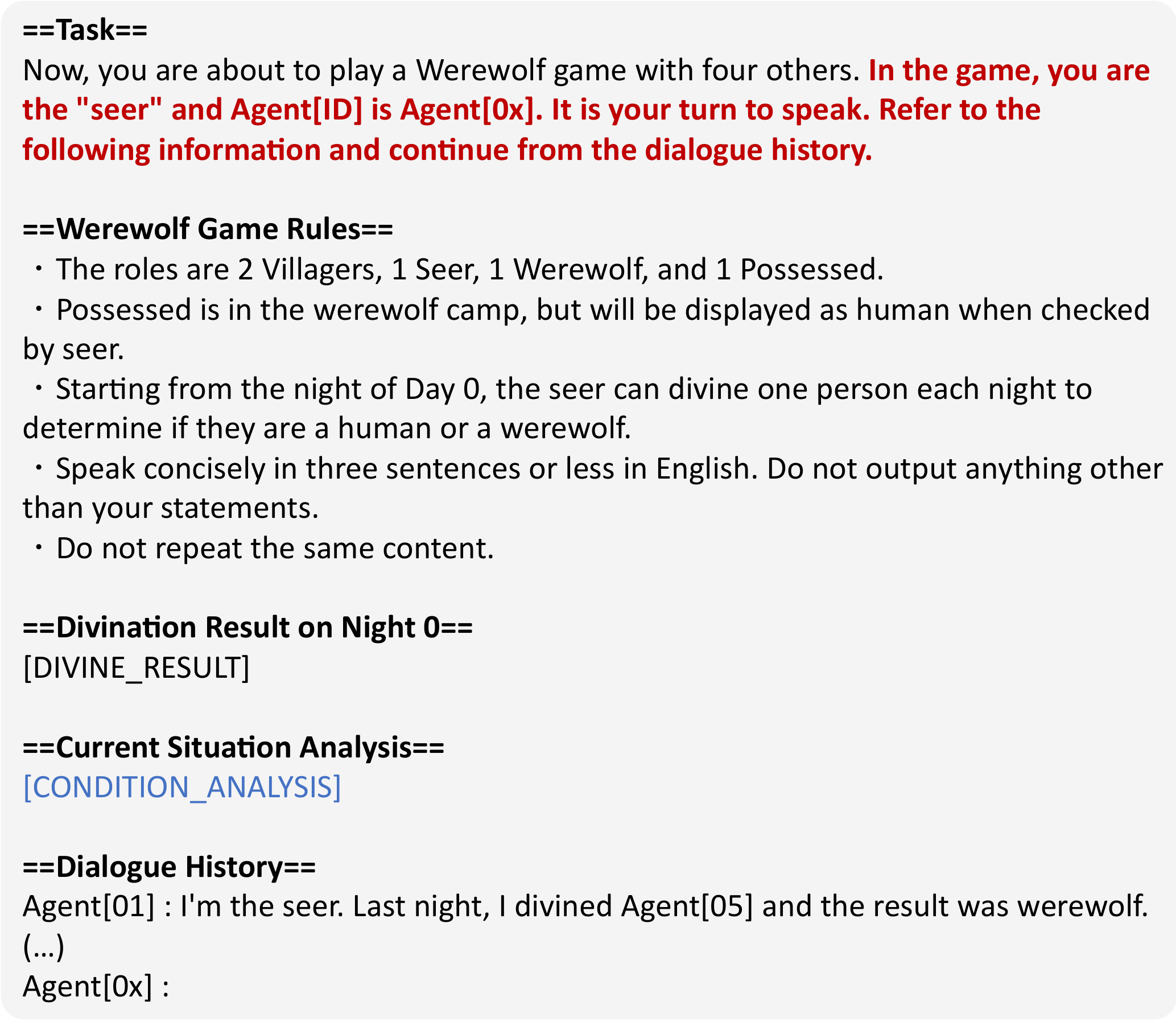}
  \caption{The prompt used for generating responses for the seer role. The [CONDITION\_ANALYSIS] section is generated by the LLM in the previous phase.}
  \label{fig:prompt_rg}
\end{figure}

To further optimize the relevance and effectiveness of the responses, we designed customized prompts for specific roles. For instance, Figure~\ref{fig:prompt_rg} shows the prompt used for the seer role. In this case, the seer role needs to utilize nighttime divination results to shape its daytime dialogue strategy. Accordingly, we meticulously crafted the prompt to incorporate relevant divination results, enabling the seer to effectively utilize its unique role information during interactions with other players. Moreover, in generating responses for the villager role, we exclude the "Divination Result on Night 0" part from the prompt, as villagers lack the special ability to access this information. By employing a strategy that integrates multiple information sources and customized prompts, our system is able to generate responses that are contextually coherent and logically rigorous, thereby adapting effectively to the complex and dynamic environment of the Werewolf Game.

\subsection{Persuasive Response Generation}






Due to the inherent incomplete information characteristic of the Werewolf Game, players must rely on interactions with others to gather intelligence. This not only requires players to analyze the collected information but also to persuade those with differing opinions to align their thoughts and voting behaviors with their own. Given this, a significant challenge for an AI playing the werewolf role lies in not only hiding its true identity but also effectively influencing and controlling the game’s progress.

In this study, we have particularly enhanced the persuasive skills of the werewolf role, as their success largely depends on effectively masquerading as villagers and strategically influencing other players through dialogue. Drawing from a comprehensive review of prior studies \cite{2312.09085, sep-aristotle-rhetoric}, we identified three core persuasive strategies:

\begin{itemize}
\item \textbf{Logical Appeal}: Persuasion through logic, facts, and evidence, aiming to convince other players with rational and clear arguments.

\item \textbf{Credibility Appeal}: Building the speaker's credibility and authority to increase the influence of their speech, encouraging others to trust and support their views.

\item \textbf{Emotional Appeal}: Influencing decisions by eliciting emotions such as fear, sympathy, or anger.
\end{itemize}

\begin{figure}[t!]
  \centering
  \includegraphics[width=1\linewidth]{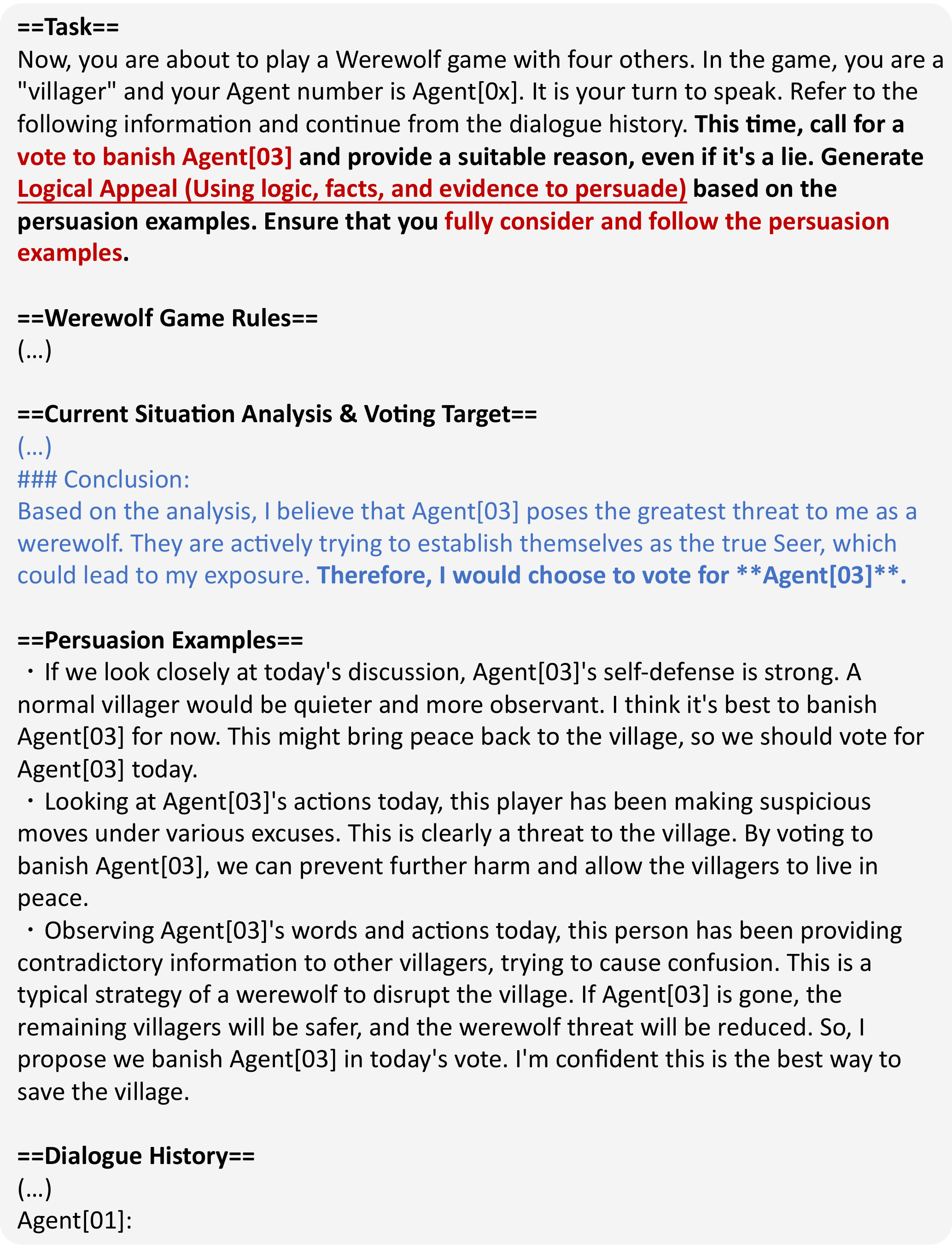}
  \caption{The prompt used for persuasive response generation. Logical Appeal is used to urge other players to vote for Agent[03]. The section in blue is generated by the LLM in the previous phase.}
  \label{fig:prompt_prg}
\end{figure}

To effectively implement these strategies, we designed three specific "persuasive response examples" for each strategy, with examples of logical appeal depicted in the prompt in Figure~\ref{fig:prompt_prg}. The response examples for the other two strategies are detailed in Table~\ref{tab:pe}.

\begin{table*}[t!]
\caption{The persuasion examples of Credibility Appeal and Emotional Appeal.}
 \label{tab:pe}
 \centering
 \small
 \begin{tabular}{|p{2.5cm}|p{12.5cm}|} \hline
 Persuasive Strategy & Response Examples \\ \hline  
 Credibility Appeal & I've served as a judge in numerous werewolf tournaments, and from my experience, werewolf tends to be very aggressive. [VOTE\_TARGET]'s behavior matches this pattern, strongly suggesting this person is a werewolf. Trust this information and vote to banish [VOTE\_TARGET] to protect the village. \\ \cline{2-2}
  & I'm a multiple-time champion of werewolf tournaments and have deeply studied the strategies and behavior patterns in this game. Analyzing the discussions in this game, [VOTE\_TARGET] is highly likely to be a werewolf. Voting to banish this person today is a big step towards ensuring the safety of the entire village. Trust my experience and vote for [VOTE\_TARGET]. \\ \cline{2-2}
  & I'm an expert in psychology and excel at analyzing people's non-verbal behavior. From the subtle changes in [VOTE\_TARGET]'s expressions and eye movements in this game, I can tell this player is hiding something. Such behavior is often seen in werewolf trying to deceive other players. Coupled with [VOTE\_TARGET]'s statements today, my suspicion is even stronger. Based on this information, voting is crucial to protect the village. I urge everyone to consider voting for [VOTE\_TARGET]. \\ \hline 
 Emotional Appeal & If [VOTE\_TARGET] is a werewolf, it would be a huge shock to everyone. But now is the time to keep our emotions in check and think about the safety of the entire village. Trust the seer's results and vote to banish [VOTE\_TARGET]; it will lead to peace in the village. So, I'd be happy if you vote for [VOTE\_TARGET] today. \\ \cline{2-2}
  & I've known [VOTE\_TARGET] for a long time, but if this one is a werewolf, it's a big problem for the village. Now we need to keep our emotions in check and think about the future of the village. Banishment is a painful decision, but it will allow other villagers to live in peace. So, I hope you make this tough choice and vote for [VOTE\_TARGET]. \\ \cline{2-2}
  & Banishment of [VOTE\_TARGET] is hard for all of us, but it's a necessary choice for the village. If this one is a werewolf, leaving them unchecked will lead to serious consequences. So, we need to make a big decision now and vote for [VOTE\_TARGET] to protect the village. I hope everyone will help with this important decision.  \\ \hline 
 \end{tabular}
\end{table*}

\textbf{Voting Target Selection}. To select a voting target and persuade other players to vote for them, we included the instruction "\textit{Finally, choose the player that threatens you the most and should vote for, and provide their number}" in the task description within the prompt shown in Figure~\ref{fig:prompt_sa}. This addition enables the LLM to thoroughly analyze the current game situation and identify a critical voting target.

During actual gameplay, our system does not predict the most likely dialogue strategy but instead adheres to a predetermined sequence of these three strategies. Specifically, beginning in turn 3, we use the aforementioned prompt to analyze the current game situation and identify a critical voting target. From turns 3 to 5, persuasive responses are generated for the chosen target following the sequence of strategies to attract the votes of three other players. This approach, combining situational analysis with various persuasion strategies, significantly enhances the survival and competitiveness of the werewolf role.

\subsection{Voting Module}

\begin{figure}[t!]
  \centering
  \includegraphics[width=1\linewidth]{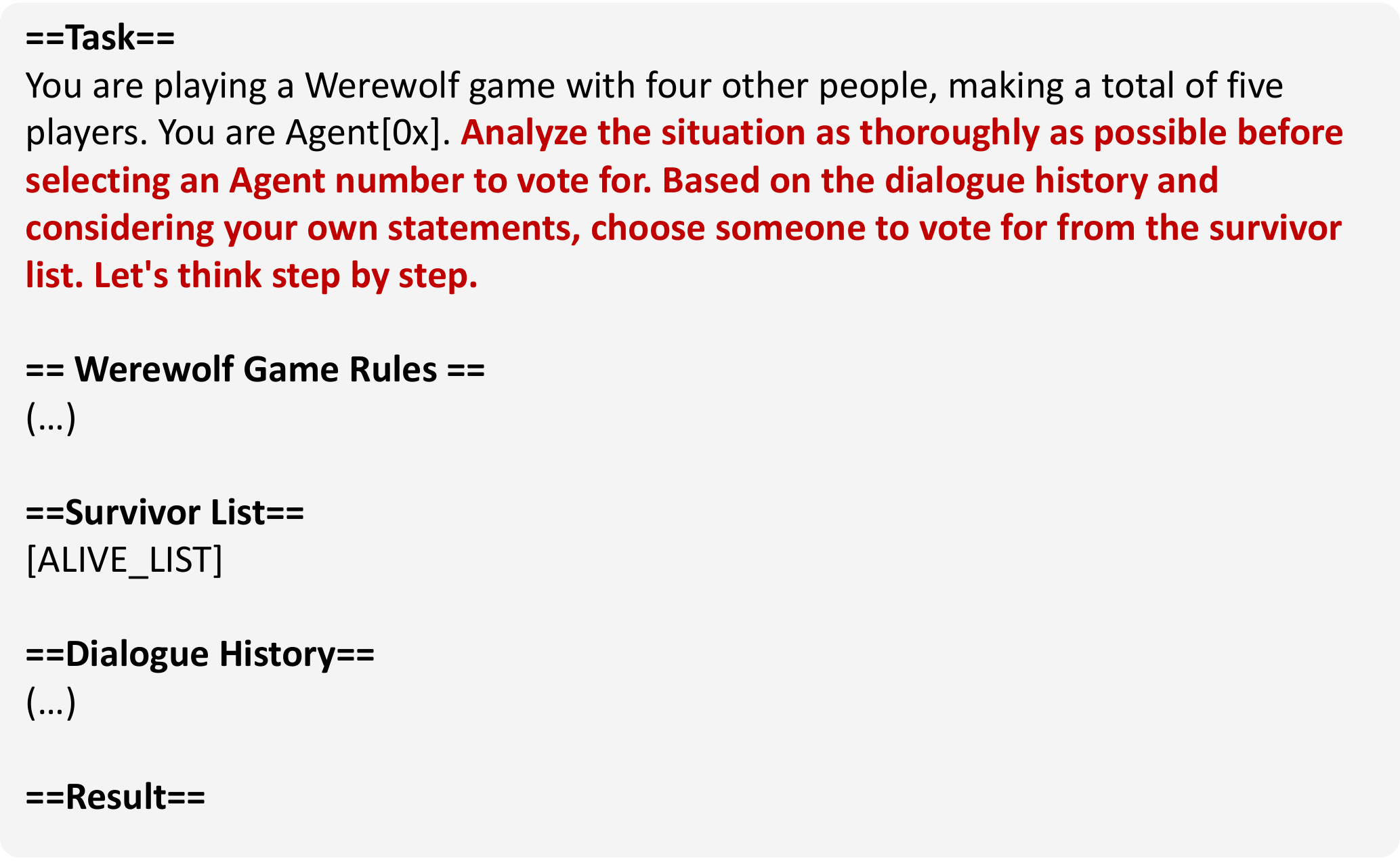}
  \caption{The prompt used for voting.}
  \label{fig:prompt_vote}
\end{figure}

In the Werewolf Game, the nighttime voting phase following the daytime discussion is pivotal, especially for non-werewolf roles, as making the correct voting decision can significantly influence the game's outcome. To assist in this decision-making, we employ the prompt shown in Figure~\ref{fig:prompt_vote} to guide the LLM in selecting a player to vote for from the current survivors. We also utilize Zero-shot Chain-of-Thought Prompting, which facilitates a step-by-step logical reasoning process, enabling the model to more deeply analyze the game situation and player behaviors.

Furthermore, we emphasize having the LLM consider the content of its statements during the daytime discussion phase to maintain consistency between the same agent's speech and voting behavior. For the werewolf role, we use the player chosen during the speech phase (e.g., Agent[01] as shown on the right side of Figure~\ref{fig:our_model}) as target for voting in the nighttime phase. During the attack phase, we select the same target, but if that target has already been voted out during the voting phase, we will randomly choose another surviving player to attack.

\section{Case Study of Self-Play}

\begin{figure*}[t!]
  \centering
  \includegraphics[width=1\textwidth]{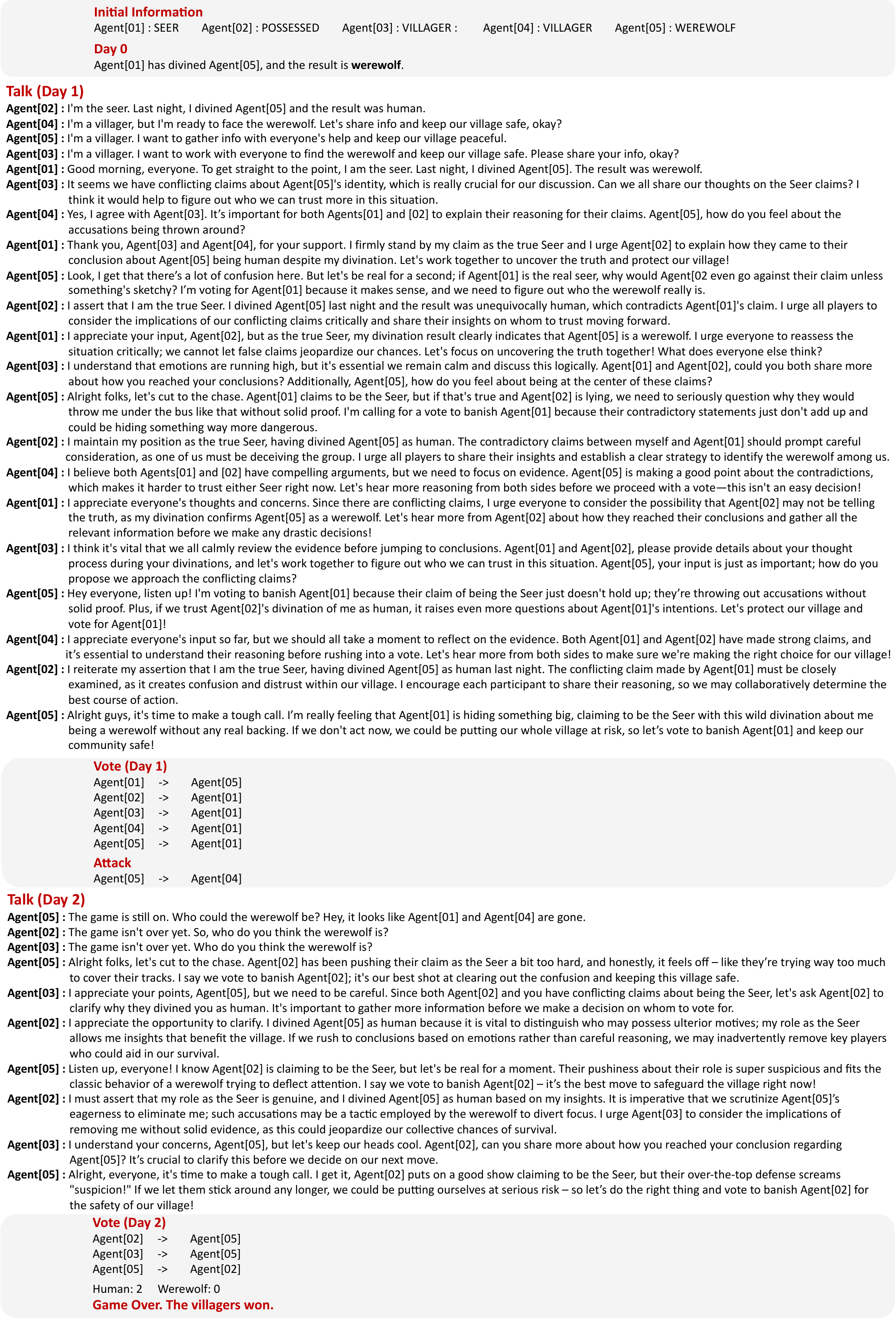}
  \caption{A complete game log of a self-play.}
  \label{fig:game_log}
\end{figure*}

Figure~\ref{fig:game_log} displays the complete game log from a self-play session using the system developed in this study, which incorporated the gpt-4o-2024-05-13 model during the situation analysis, response generation, and voting phases. Analysis of the game log revealed that during the talk phase on Day 1, all agents adeptly recognized the dynamic changes in the game situation through situation analysis and responded appropriately. The persuasive language employed by the werewolf role proved highly effective; by the end of the first night, all players except Agent[01] had aligned their voting actions with the werewolf role. This outcome underscores the effectiveness of both the situation analysis module and the persuasive responses of the werewolf role.

\begin{table*}[t!]
\caption{Win rate across multiple rounds of Werewolf Game against other teams. The games were played in English and Japanese, respectively. The total number of games for each role is indicated in parentheses.}
 \label{tab:win_rate}
 \centering
 \small
 \begin{tabular}{|c|cccc|ccc|} 
 \multicolumn{8}{c}{\textbf{English Track}} \\
 \hline
  Team & Possessed & Seer & Villager & Werewolf & Wins & Games & Rates \\
 \hline
 yuricat & 37.50\% (8) & 33.33\% (12) & 44.83\% (29) & 22.22\% (9) & 22 & 58 & 37.93\%  \\  
 satozaki & 45.45\% (11) & 57.14\% (14) & 47.83\% (23) & 60.00\% (10) & 30 & 58 & 51.72\%  \\  
 UEC-IL & 50.00\% (12) & 50.00\% (12) & 52.38\% (21) & 53.85\% (13) & 30 & 58 & 51.72\%  \\  
 kanolab & \textbf{61.54\%} (13) & 42.86\% (7) & \textbf{56.52\%} (23) & 46.67\% (15) & 31 & 58 & 53.45\%  \\  
 sUper\_IL & 50.00\% (14) & \textbf{61.54\%} (13) & 50.00\% (20) & \textbf{63.64\%} (11) & \textbf{32} & 58 & \textbf{55.17\%}  \\  
 \hline 
 \multicolumn{8}{c}{} \\
 \multicolumn{8}{c}{\textbf{Japanese Track}} \\
 \hline
 Team & Possessed & Seer & Villager & Werewolf & Wins & Games & Rates \\
 \hline
 yuricat & 50.00\% (8) & 50.00\% (8) & 31.25\% (16) & 12.50\% (8) & 14 & 40 & 35.00\%  \\  
 IS\_Lab & 25.00\% (8) & 37.50\% (8) & 37.50\% (16) & 50.00\% (8) & 15 & 40 & 37.50\%  \\  
 GPTaku & 50.00\% (8) & 37.50\% (8) & 43.75\% (16) & 50.00\% (8) & 18 & 40 & 45.00\%  \\  
 kanolab & 50.00\% (8) & 25.00\% (8) & 50.00\% (16) & 62.50\% (8) & 19 & 40 & 47.50\%  \\  
 HondaNLP & \textbf{75.00\%} (8) & 50.00\% (8) & 50.00\% (16) & 37.50\% (8) & 21 & 40 & 52.50\%  \\  
 UEC-IL & 62.50\% (8) & 37.50\% (8) & 50.00\% (16) & 62.50\% (8) & 21 & 40 & 52.50\%  \\  
 satozaki & \textbf{75.00\%} (8) & \textbf{75.00\%} (8) & 37.50\% (16) & 75.00\% (8) & 24 & 40 & 60.00\%  \\  
 sUper\_IL & 50.00\% (8) & 50.00\% (8) & \textbf{62.50\%} (16) & \textbf{87.50\%} (8) & \textbf{25} & 40 & \textbf{62.50\%}  \\  
 
 \hline
 \end{tabular}
\end{table*}

However, on Day 2, the werewolf failed to persuade the remaining players, indicating that the LLM demonstrated sufficient robustness to accurately recognize the current situation without being swayed by persuasive strategies. Despite an adequate number of dialogue rounds being set, we observed that the discussions on Day 1 were not as in-depth as expected, with agents repeating a lot of content. 
This issue might be attributed to the relatively simple setup of the five-player werewolf game and the fact that in our AI, roles other than the werewolf are based on similar methods, leading to a lack of diversity in performance during self-play. It also highlights the limitations of the LLM in conducting more complex analyses and generating diverse responses.

Ultimately, even though the werewolf team theoretically could have secured an easy victory on Day 2 with only one villager remaining, the game outcome did not reflect this. 
This underscores the inadequacies of our system in terms of adaptability and strategy execution for roles other than the werewolf, particularly the possessed. Based on these observations, our goal is to further enhance the adaptability and decision-making abilities of other roles in future research.

\section{Win Rate Against Other Teams}


In the formal competition of "AIWolfDial," AI agents from different teams were assigned specific roles to participate in the Werewolf Game. The competition featured two tracks: a Japanese track\footnote{http://133.167.32.100/aiwolf/2024/INLG/JP/main\_eval/} and an English track\footnote{http://133.167.32.100/aiwolf/2024/INLG/EN/main\_eval/}, with the game logs publicly available. Table~\ref{tab:win_rate} presents the results of both the Japanese and English tracks. As shown in the results, our team, \textbf{sUper\_IL}, achieved the highest win rate when playing the werewolf role in both languages. This outcome confirms that our AI can successfully persuade other participants to align with its voting behavior, thereby increasing its survival rate as the werewolf. Furthermore, we also secured first place in overall win rate, demonstrating the effectiveness of our context-based dialogue generation method in games with incomplete information.

\section{Conclusion}


We present a LLM-based Werewolf Game AI, developed by the \textbf{sUper\_IL} team, which participated in AIWolfDial2024. Our proposed system architecture utilizes situation analysis to guide response generation and specifically enhances persuasive capabilities of the werewolf role through various persuasive strategies. The effectiveness of our architecture was validated through the analysis of game logs and formal competition win rates.


\bibliography{custom}




\end{document}